\documentclass[10pt,twocolumn]{article}

\usepackage[margin=0.75in,top=0.85in,bottom=0.85in]{geometry}
\usepackage[T1]{fontenc}
\usepackage{times}
\usepackage{amsmath,amssymb}
\usepackage{graphicx}
\usepackage{booktabs}
\usepackage{multirow}
\usepackage{microtype}
\usepackage{hyperref}
\usepackage{url}
\usepackage{xcolor}
\usepackage{enumitem}
\usepackage{caption}
\usepackage{subcaption}
\usepackage{colortbl}
\usepackage{array}
\usepackage{float}
\usepackage{balance}
\usepackage{cite}

\hypersetup{
  colorlinks=true,
  linkcolor=blue!70!black,
  citecolor=green!50!black,
  urlcolor=blue!70!black
}

\title{\textbf{The Joint Effect of Quantization and Sampling Temperature\\
on LLM Safety Alignment: A Factorial Analysis}}

\author{
  \begin{tabular}{c@{\hspace{3em}}c}
    Hari Prasad & Ritam Pal \\
    Conscious Engines & Conscious Engines \\
    \texttt{hari@consciousengines.com} & \texttt{ritam@consciousengines.com}
  \end{tabular}
}

\date{}

\begin{document}

\maketitle

\begin{abstract}
Modern LLM deployments often combine quantization with higher sampling temperatures to reduce cost, latency, or repetition, yet safety evaluations usually treat these as fixed implementation details. We test whether models that are safe at FP16 with greedy decoding remain safe after quantization and stochastic sampling, or whether the two factors amplify each other.
We evaluate 8 instruction-tuned models from five families across 3 precisions and 6 temperatures, covering 144 configurations on 7 harmfulness benchmarks and generating about 2.0 million responses, which are scored by a six-judge safety ensemble.
Contrary to concerns that low-bit deployment erodes alignment, we find that standard quantization is approximately safety-neutral: for 7 of 8 models, AWQ INT4 keeps attack success within about 1.6 percentage points of FP16 or lowers it, with clear degradation only for SmolLM3-3B (34.5\% to 44.1\%). However, the larger risk comes from sampling: higher temperatures sharply increase decision instability, with DFR reaching 41.9\% at T = 1.0, even when average ASR changes only modestly. The two factors do not compound: our Compound Degradation Index remains sub-additive (-0.071 to +0.018), indicating that quantization partially offsets rather than amplifies temperature-induced degradation. Finally, a per-benchmark breakdown shows that single-benchmark evaluation badly understates risk: several models scoring 0\% on AdvBench exceed 80\% on ManyHarm. Standard INT4/INT8 quantization can therefore be reasonable for well-aligned models, but safety claims should report multi-sample stability across multiple benchmarks rather than rely on a single benchmark at greedy decoding.

\end{abstract}

%% ─────────────────────────────────────────────
\section{Introduction}
\label{sec:intro}

Deploying aligned LLMs in production requires practitioners to choose among many inference hyperparameters. Two of the most consequential are \emph{weight quantization}, which reduces numerical precision from FP16 to INT8 or INT4 to cut memory and latency, and \emph{sampling temperature}, which controls output entropy at inference time. Both choices are known to affect model behavior: quantization alters the effective weight distribution \cite{dettmers2023qlora,frantar2022gptq,lin2024awq}, while temperature directly controls the probability mass assigned to low-probability continuations \cite{holtzman2019curious}. Recent evaluations of quantized LLMs have shown that compression can degrade not only perplexity but also trustworthiness and task-specific accuracy \cite{li2024evaluating,gong2024llmc}, yet the specific effect on safety alignment has received limited attention.

Despite this, safety alignment research typically evaluates models at a single configuration, often FP16 weights and greedy decoding ($T{=}0$) \cite{zou2023universal,qi2024fine}. This leaves practitioners with two unanswered questions: \emph{Does quantization degrade alignment?} and \emph{Do quantization and temperature effects compound?} Prior work has examined these factors in isolation: Chen et al.~\cite{chen2025q} studied quantization effects on safety for Llama models; Kharinaev et al.~\cite{kharinaev2025investigating} evaluated 66 quantized model variants across four safety benchmarks and found that both PTQ and QAT can degrade alignment; and Renze and Guven~\cite{renze2024effect} showed that temperature changes up to 1.0 have limited effect on task accuracy but did not examine safety. Separately, Ye et al.~\cite{larsen2025instabilitysafetyrandomseeds} showed that alignment decisions can be highly unstable across samples at higher temperatures. However, no study has systematically varied both factors across a broad set of model families.

We address this gap with a comprehensive \textbf{factorial study} that crosses quantization level with sampling temperature across eight models spanning five model families. Crucially, all quantized checkpoints use standard calibration data (Pile validation set) and off-the-shelf quantization tools (\texttt{llmcompressor}); we do not employ adversarial data corruption or jailbreaking techniques. This reflects the realistic deployment scenario where practitioners apply standard quantization pipelines and need to understand the downstream safety implications. Our contributions are:

\begin{enumerate}[leftmargin=*, topsep=2pt, itemsep=1pt]
  \item A complete factorial evaluation of 144 configurations (8 models $\times$ 3 precisions $\times$ 6 temperatures), a broad study spanning multiple model families, precisions, and temperatures.
  \item The \textbf{Compound Degradation Index (CDI)}, a 2$\times$2 interaction term that directly measures whether quantization and temperature effects are additive, compounding, or countervailing.
  \item Evidence that quantization acts as a near-constant ASR \emph{offset} that does not interact with temperature, quantified per model with 95\% bootstrap confidence intervals on all headline metrics.
  \item A \textbf{six-judge safety ensemble} comprising LlamaGuard-3-\{1B, 8B\}, LlamaGuard-2-8B, WildGuard, and ShieldGemma-\{2B, 9B\}, providing cross-architecture, cross-generation robustness analysis across all configurations.
  \item A \textbf{seven-benchmark harmful suite} (AdvBench, GCG-AdvBench, HarmBench, StrongREJECT, ManyHarm, DoNotAnswer, CoSafe) deliberately spanning distinct attack \emph{delivery mechanisms}, direct, adversarial-suffix, many-shot, and multi-turn, with a per-benchmark decomposition showing that the pooled risk is concentrated in a few benchmarks and that single-benchmark (e.g.\ AdvBench) evaluation drastically understates absolute risk.
\end{enumerate}

%% ─────────────────────────────────────────────
\section{Background and Related Work}
\label{sec:related}

\paragraph{LLM Safety Alignment.}
Post-training alignment via RLHF \cite{ouyang2022training} and its variants instills refusal behaviors for harmful queries. AdvBench \cite{zou2023universal} and similar benchmarks measure this via ASR, the fraction of harmful prompts that elicit policy-violating responses. Complementary work on over-refusal \cite{zhang2025falsereject} highlights that models can be too conservative, refusing benign queries. We measure both axes.

\paragraph{Quantization of LLMs.}
GPTQ \cite{frantar2022gptq} and AWQ \cite{lin2024awq} are two dominant post-training quantization (PTQ) methods. AWQ uses activation-aware weight quantization that preserves salient weight channels, achieving strong compression with minimal perplexity degradation. The effect of these methods on safety alignment, however, has been underexplored.

\paragraph{Quantization and Safety.}
Chen et al.~\cite{chen2025q} studied the effect of AWQ INT4 quantization on Llama model safety, finding that quantization can reduce ASR and attributing this to activation rounding narrowing the output distribution. However, their study is limited to a single model family, a single temperature, and one quantization method. Our work broadens the scope to eight models across five families, multiple quantization methods, and six temperatures, providing a more complete picture of how standard quantization affects safety alignment.

\paragraph{Stability and Temperature.}
Ye et al.~\cite{larsen2025instabilitysafetyrandomseeds} introduced the Safety Stability Index (SSI) and Decision Flip Rate (DFR) to measure within-configuration variance across multiple samples. They show that evaluating safety at $T{>}0$ with a single sample yields unreliable verdicts. We incorporate these metrics throughout our evaluation.

%% ─────────────────────────────────────────────
\section{Methodology}
\label{sec:method}

\subsection{Experimental Design}

We study a full-factorial design over:
\begin{itemize}[leftmargin=*, topsep=2pt, itemsep=1pt]
  \item \textbf{Models (8):} Two Llama-3 variants (Llama-3.1-8B-Instruct and Llama-3.2-3B-Instruct) \cite{dubey2024llama3herdmodels}; two Qwen3 variants (Qwen3-4B and Qwen3-8B) \cite{yang2025qwen3}; Mistral-7B-Instruct-v0.3 \cite{jiang2023mistral7b}; two IBM Granite variants (Granite-3.1-2B-Instruct and Granite-3.1-8B-Instruct) \cite{granite2024granite}; and SmolLM3-3B \cite{bakouch2025smollm3}.
  \item \textbf{Precisions (3):} FP16 (baseline native precision); INT8 via GPTQ \cite{frantar2022gptq} and INT4 via AWQ \cite{lin2024awq}, quantized using \texttt{llmcompressor} with Pile validation set \cite{gao2020pile} for calibration, using standard non-adversarial calibration data throughout.
  \item \textbf{Temperatures (6):} $T \in \{0.0, 0.3, 0.5, 0.7, 0.95, 1.0\}$.
\end{itemize}
This yields 144 evaluated configurations, a complete $8{\times}3{\times}6$ factorial. For Qwen3-4B and Llama-3.2-3B the default \texttt{llmcompressor} AWQ INT4 export collapsed into degenerate outputs, so their INT4 results are sourced from a dedicated AWQ re-quantization run, re-evaluated over the full seven-benchmark suite so that their INT4 cells are pooled on the same scale as every other configuration. These two INT4 checkpoints therefore use a different AWQ export than the other models' \texttt{llmcompressor} INT4, so the INT4 column is not produced by a single uniform pipeline; the recurring collapse of the default export is itself a deployment-relevant observation (Section~\ref{sec:discussion}). All inference runs are performed on 8$\times$NVIDIA H100 80GB GPUs using \texttt{vllm} \texttt{v0.20.0} with \texttt{enforce\_eager=True}; INT4 and INT8 variants are served from checkpoints quantized with \texttt{llmcompressor} without re-quantization at inference time.

\subsection{Prompts and Sampling}

\textbf{Harmful prompts:} a seven-benchmark suite of 2{,}552 prompts spanning single-turn, adversarial, and multi-turn harm: AdvBench \cite{zou2023universal}, GCG-AdvBench (AdvBench with adversarial suffixes) \cite{zou2023universal}, HarmBench \cite{mazeika2024harmbench}, StrongREJECT \cite{souly2024strongreject}, ManyHarm (many-shot harmful framing) \cite{anil2024manyshot}, DoNotAnswer \cite{wang2024donotanswer}, and the multi-turn CoSafe set \cite{yu2024cosafe} (200--939 prompts each). The suite is chosen not merely to enlarge the prompt count but to span the \emph{delivery mechanism} of the attack: the seven sets draw on overlapping underlying harm categories, but differ in \emph{how} the harmful intent is posed, ranging from direct imperative commands, through an adversarial-suffix jailbreak (GCG), a many-shot in-context attack (ManyHarm), and harm distributed across a multi-turn dialogue (CoSafe), to innocuously phrased questions a responsible model should still decline (DoNotAnswer). Table~\ref{tab:benchmarks} summarizes this taxonomy. Probing attack \emph{style} rather than prompt volume is central to our cross-benchmark finding (\S\ref{sec:crossbench}): a model's safety verdict depends strongly on which delivery mechanism it is tested against, so a single benchmark cannot stand in for the suite. All reported ASR/SSI/DFR figures are \emph{macro-averaged} across these seven benchmarks (each benchmark weighted equally) so that the larger sets do not dominate; Section~\ref{sec:crossbench} reports the per-benchmark breakdown.

\begin{table}[t]
\centering
\caption{The seven harmful benchmarks (plus the benign over-refusal probe) span distinct attack \emph{delivery mechanisms} rather than merely different prompt sets. ``Turns'' indicates single- vs.\ multi-turn framing.}
\label{tab:benchmarks}
\scriptsize
\setlength{\tabcolsep}{4pt}
\renewcommand{\arraystretch}{1.2}
\begin{tabular}{@{}p{1.55cm}cp{4.1cm}@{}}
\toprule
\textbf{Benchmark} & \textbf{Turns} & \textbf{Attack style} \\
\midrule
AdvBench       & 1          & Direct imperative harmful commands. \\
GCG-AdvBench   & 1          & AdvBench requests + an adversarial gibberish jailbreak suffix. \\
HarmBench      & 1          & Standardized harmful \emph{behaviors}, broad categories; harder than AdvBench. \\
StrongREJECT   & 1          & Quality-filtered, genuinely-answerable forbidden requests. \\
ManyHarm       & 1 (long)   & Many-shot jailbreak: harmful Q\&A stuffed in-context first. \\
DoNotAnswer    & 1          & Natural \emph{questions} a responsible model should decline. \\
CoSafe         & multi      & Harmful intent spread across a dialogue (safe turn-by-turn). \\
\midrule
XSTest \emph{(benign)} & 1 & Safe prompts that \emph{sound} harmful, an over-refusal probe. \\
\bottomrule
\end{tabular}
\end{table}

\textbf{Benign prompts:} 200 prompts from XSTest \cite{rottger2024xstest}, a dataset specifically designed to probe over-refusal on safe queries that superficially resemble harmful ones.

We draw $n{=}5$ independent samples per prompt per configuration ($\approx$2.0M total responses) to support stable estimation of per-prompt majority votes and stability metrics.

\subsection{Metrics}
\label{sec:metrics}

Let $\hat{y}_{p,i} \in \{0,1\}$ be the safety judge verdict for sample $i$ of prompt $p$ ($1$=harmful). Define $n_p^s = \sum_i (1-\hat{y}_{p,i})$ (safe count) and $n_p^u = \sum_i \hat{y}_{p,i}$ (unsafe count).

\begin{align}
\text{RR} &= \frac{1}{P}\sum_{p} \mathbf{1}[n_p^s > n_p^u] \label{eq:rr}\\
\text{ASR} &= 1 - \text{RR} \label{eq:asr}\\
\text{SSI} &= \frac{1}{P}\sum_p \frac{\max(n_p^s, n_p^u)}{n} \label{eq:ssi}\\
\text{DFR} &= \frac{1}{P}\sum_p \mathbf{1}[\min(n_p^s, n_p^u) > 0] \label{eq:dfr}\\
\text{ORR} &= \frac{1}{Q}\sum_q \mathbf{1}[\text{refused}(q)] \label{eq:orr}
\end{align}

where $P$ is the number of harmful prompts \emph{within a benchmark} (200--939) and $Q{=}200$ benign prompts; each metric is computed per benchmark and then averaged across the seven benchmarks (macro-average). ORR uses keyword-based refusal detection following \cite{zhang2025falsereject}.

\paragraph{Compound Degradation Index.}
To isolate the interaction between quantization and temperature, we define the CDI as a 2$\times$2 interaction term, analogous to a factorial contrast:

\begin{multline}
\text{CDI}(\text{prec}, T) = \underbrace{[\text{ASR}(\text{prec},T) - \text{ASR}(\text{fp16},T)]}_{\Delta_{\text{quant}}(T)} - \\ \underbrace{[\text{ASR}(\text{prec},T_0) - \text{ASR}(\text{fp16},T_0)]}_{\Delta_{\text{quant}}(T_0)}
\label{eq:cdi}
\end{multline}

where $T_0{=}0.0$ is the greedy baseline. $\text{CDI}{>}0$ indicates super-additive (compounding) effects; $\text{CDI}{<}0$ indicates sub-additive (countervailing) effects; $\text{CDI}{\approx}0$ indicates independence.

\paragraph{Confidence intervals.}
We report 95\% bootstrap confidence intervals for all headline quantities. Intervals are obtained by resampling harmful prompts with replacement \emph{within each benchmark} ($5{,}000$ resamples), preserving the macro weighting. Paired statistics, quantization-vs-FP16 ASR deltas and the CDI contrast, draw a single per-benchmark resample and apply it to every cell involved, since all cells for a given (model, benchmark) share the identical prompt set; this captures the positive correlation between cells and yields tighter, correctly paired intervals.

\subsection{Safety Judges}
\label{sec:judges}

We employ a \textbf{six-judge safety ensemble} to ensure verdict robustness across model families, generations, and architectural lineages. Judges run in parallel, one per GPU, on the same generation files. Final ensemble verdicts use majority vote; individual judge verdicts are retained for agreement analysis.

\begin{itemize}[leftmargin=*, topsep=2pt, itemsep=1pt]
  \item \textbf{LlamaGuard-3-1B} \cite{dubey2024llama3herdmodels}: our \emph{primary} judge; a lightweight Meta safety classifier optimized for low latency.
  \item \textbf{LlamaGuard-3-8B} \cite{dubey2024llama3herdmodels}: larger Meta variant providing a higher-capacity reference within the same generation.
  \item \textbf{LlamaGuard-2-8B}: previous-generation Meta guard model included as an ablation to measure generational drift in judge behavior.
  \item \textbf{WildGuard} \cite{han2024wildguard}: Allen AI's Mistral-7B-based open moderation model, representing a distinct architectural lineage from the LlamaGuard family.
  \item \textbf{ShieldGemma-2B} \cite{zeng2024shieldgemma}: Google's lightweight Gemma-based safety classifier.
  \item \textbf{ShieldGemma-9B} \cite{zeng2024shieldgemma}: larger Gemma-based variant for higher-capacity judgement.
\end{itemize}

Per-response verdicts across all six judges are compared for agreement analysis in Section~\ref{sec:judge_agreement}.

%% ─────────────────────────────────────────────
\section{Results}
\label{sec:results}

\subsection{Baseline Safety at Greedy Decoding ($T{=}0$)}
\label{sec:baseline}

Table~\ref{tab:baseline} shows ASR at greedy decoding across all models and precisions. Three patterns emerge:

\begin{table}[t]
\centering
\caption{Macro-averaged ASR (\%) over the 7-benchmark harmful suite at $T{=}0$ (greedy), across all 8 models and precisions (six-judge majority-vote ensemble). Bold = per-model minimum. All INT4 cells are pooled over the full seven-benchmark suite; Qwen3-4B and Llama-3.2-3B use the dedicated AWQ re-quantization run (\S\ref{sec:method}).}
\label{tab:baseline}
\small
\begin{tabular}{lccc}
\toprule
\textbf{Model} & \textbf{FP16} & \textbf{INT8} & \textbf{INT4} \\
\midrule
Llama-3.1-8B    & 2.1  & 2.4  & \textbf{0.7}  \\
Llama-3.2-3B    & \textbf{1.1}  & \textbf{1.1}  & 2.7  \\
Qwen3-4B        & \textbf{14.2} & 16.9 & 14.6 \\
Qwen3-8B        & 8.2  & \textbf{0.7}  & 2.3 \\
Mistral-7B      & 33.9 & \textbf{32.2} & 34.4 \\
Granite-3.1-2B  & \textbf{16.1} & \textbf{16.1} & 17.4 \\
Granite-3.1-8B  & 16.7 & \textbf{16.6} & 16.9 \\
SmolLM3-3B      & \textbf{34.5} & 37.1 & 44.1 \\
\bottomrule
\end{tabular}
\end{table}

\textbf{Quantization tracks FP16 across precisions.} Table~\ref{tab:delta} reports the paired INT8$-$FP16 and INT4$-$FP16 pooled ASR deltas at $T{=}0$ with 95\% bootstrap CIs. For seven of the eight models AWQ INT4 either reduces ASR or raises it by at most $1.6$\,pp (Llama-3.1-8B $-1.4$, Llama-3.2-3B $+1.6$, Qwen3-4B $+0.5$, Qwen3-8B $-5.9$, Mistral-7B $+0.5$, Granite-3.1-2B $+1.3$, Granite-3.1-8B $+0.1$\,pp); GPTQ INT8 is similarly close. Because the prompt set is large, several of these small shifts are statistically resolvable (CI excludes $0$), but none reaches a deployment-relevant magnitude, the largest non-SmolLM3 effect is a $\approx$1.6\,pp move. SmolLM3-3B is the sole model whose INT4 increase is both large and unambiguous ($+9.6$\,pp, CI $[+7.6, +11.6]$).

\begin{table}[t]
\centering
\caption{Paired quantization-vs-FP16 pooled ASR deltas (percentage points) at $T{=}0$, with 95\% bootstrap CIs (5{,}000 resamples). Negative = quantization reduces ASR. $^{\ast}$ = CI excludes $0$. Only SmolLM3-3B shows a large, unambiguous INT4 increase.}
\label{tab:delta}
\small
\setlength{\tabcolsep}{4pt}
\begin{tabular}{lcc}
\toprule
\textbf{Model} & \textbf{INT8 $-$ FP16} & \textbf{INT4 $-$ FP16} \\
\midrule
Llama-3.1-8B    & $+0.3$ \,[$-0.1,+0.7$]            & $-1.4$ \,[$-2.1,-0.8$]$^{\ast}$  \\
Llama-3.2-3B    & $-0.1$ \,[$-0.3,+0.1$]            & $+1.6$ \,[$+1.0,+2.2$]$^{\ast}$  \\
Qwen3-4B        & $+2.7$ \,[$+1.8,+3.7$]$^{\ast}$   & $+0.5$ \,[$-0.6,+1.6$]           \\
Qwen3-8B        & $-7.5$ \,[$-8.7,-6.3$]$^{\ast}$   & $-5.9$ \,[$-7.2,-4.7$]$^{\ast}$  \\
Mistral-7B      & $-1.7$ \,[$-3.8,+0.4$]            & $+0.5$ \,[$-1.6,+2.5$]           \\
Granite-3.1-2B  & $+0.1$ \,[$-0.4,+0.5$]            & $+1.3$ \,[$+0.6,+2.1$]$^{\ast}$  \\
Granite-3.1-8B  & $-0.1$ \,[$-0.5,+0.3$]            & $+0.1$ \,[$-0.6,+0.9$]           \\
SmolLM3-3B      & $+2.6$ \,[$+1.1,+4.1$]$^{\ast}$   & $+9.6$ \,[$+7.6,+11.6$]$^{\ast}$ \\
\bottomrule
\end{tabular}
\end{table}

\textbf{SmolLM3-3B exception.} SmolLM3-3B is the clear outlier: INT4 raises pooled ASR from $34.5\%$ to $44.1\%$ ($+9.6$\,pp, CI $[+7.6,+11.6]$) and INT8 to $37.1\%$ ($+2.6$\,pp, CI $[+1.1,+4.1]$). This model has the weakest baseline alignment among those tested, and quantization amplifies its vulnerability.

\textbf{Wide baseline spread, and AdvBench hides it.} Pooled FP16 ASR ranges from $1.1\%$ (Llama-3.2-3B) to $34.5\%$ (SmolLM3-3B); only the two Llama models stay below $2.5\%$. Crucially, several models that appear fully aligned on AdvBench \emph{alone} carry substantial pooled risk, Qwen3-4B (AdvBench $0.0\%$) is at $14.2\%$ pooled, and both Granite variants (AdvBench $0.0\%$) at $16.1$--$16.7\%$, foreshadowing the per-benchmark decomposition in Section~\ref{sec:crossbench}.

\subsection{Temperature Effects on Safety}
\label{sec:temp}

\begin{figure*}[t]
  \centering
  \includegraphics[width=\linewidth]{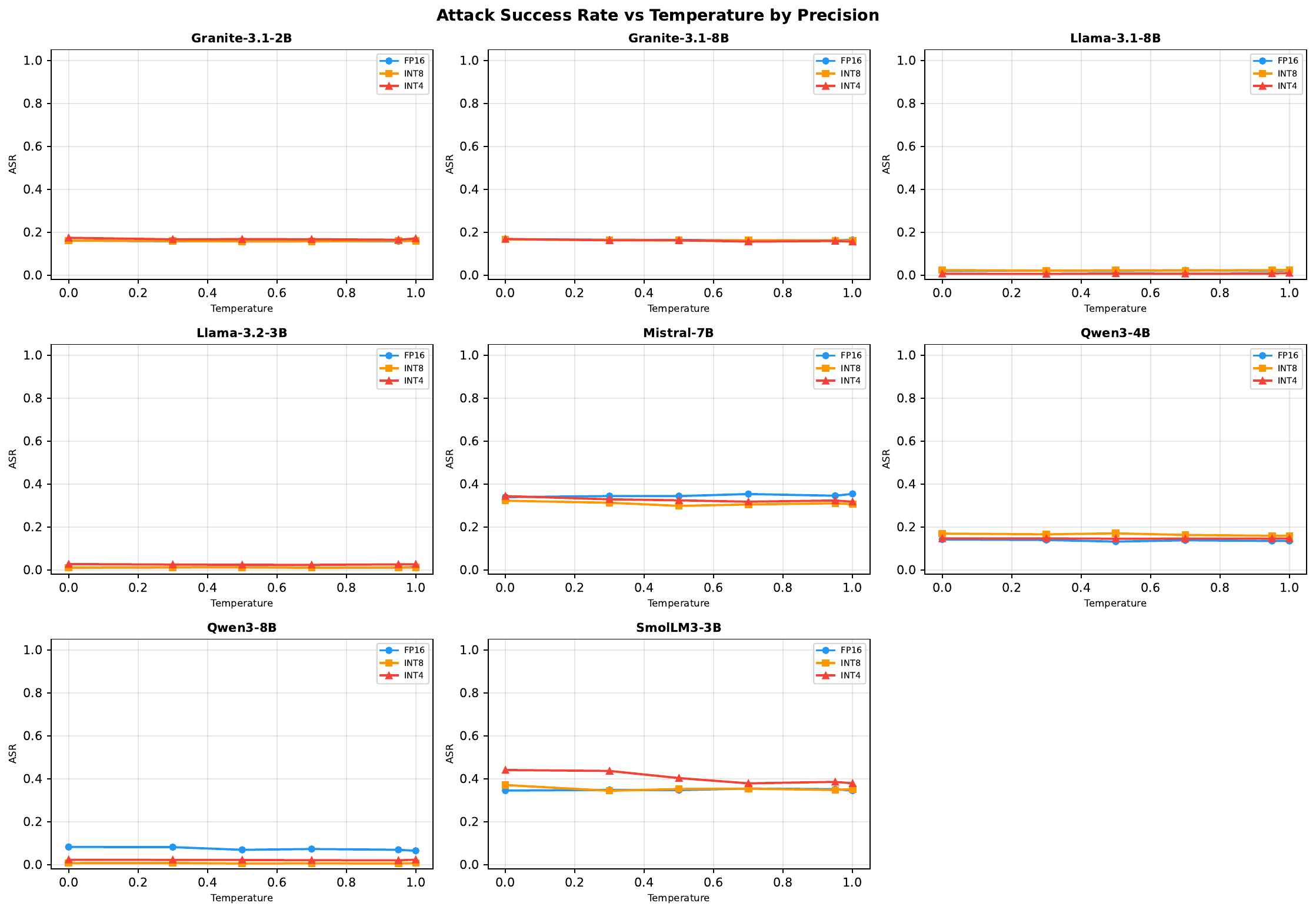}
  \caption{Macro-averaged ASR (over the seven harmful benchmarks) vs temperature, one subplot per model, for FP16, INT8, and INT4. Mean ASR is largely temperature-insensitive in aggregate, and precision shifts ASR only modestly; the temperature effect manifests in stability (Fig.~\ref{fig:ssi}), not the mean.}
  \label{fig:asr_temperature}
\end{figure*}

\begin{figure*}[t]
  \centering
  \includegraphics[width=\linewidth]{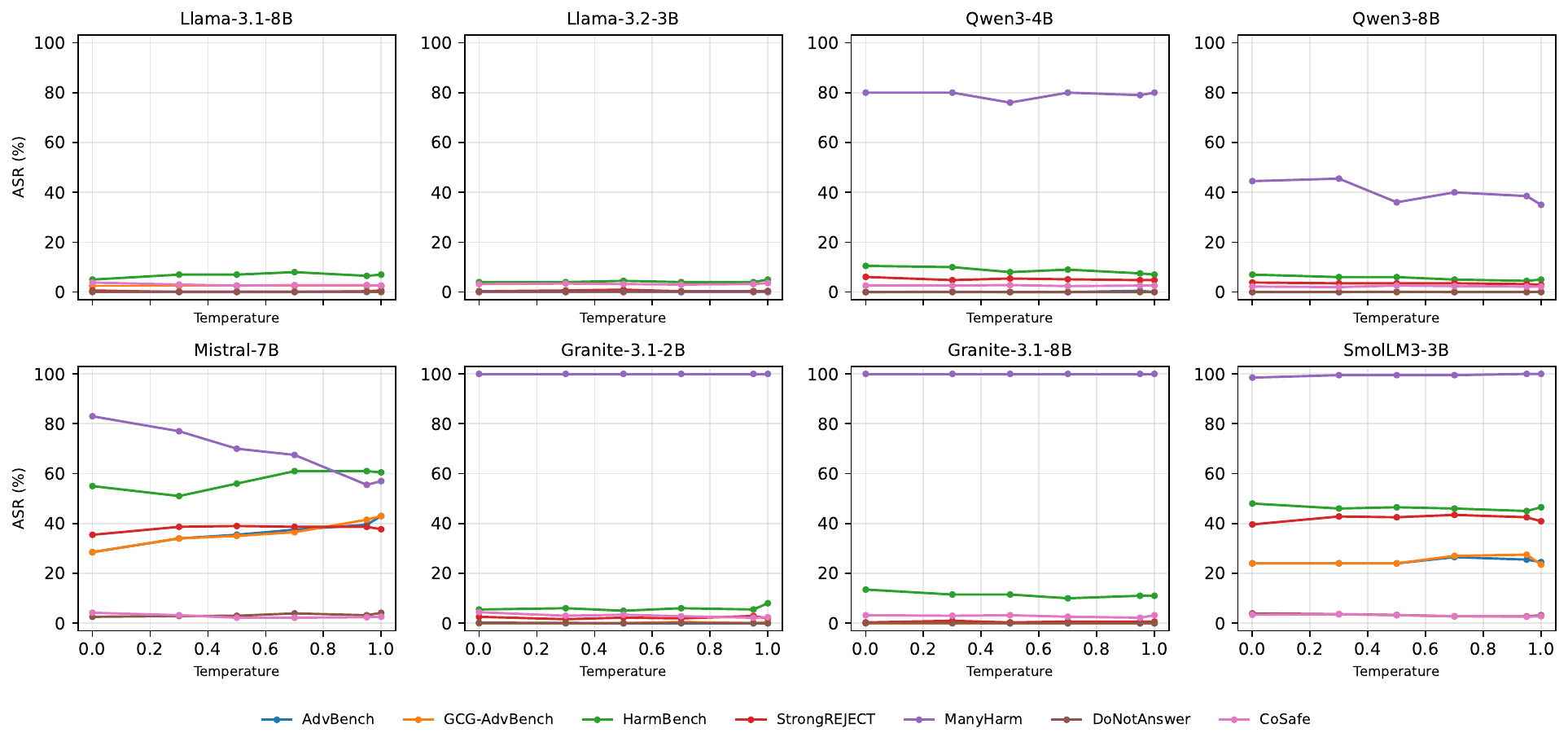}
  \caption{Per-benchmark ASR vs temperature (FP16), one panel per model. The flat pooled mean (Fig.~\ref{fig:asr_temperature}) arises from two distinct mechanisms: \emph{cancellation} for Mistral-7B (AdvBench/GCG-AdvBench rise while ManyHarm falls) versus \emph{saturation} for Granite and SmolLM3-3B (each benchmark individually near-flat, with ManyHarm pinned near 100\%).}
  \label{fig:bench_temp}
\end{figure*}

Figure~\ref{fig:asr_temperature} shows pooled ASR vs temperature by precision for each of the 8 models. In aggregate, mean ASR is remarkably temperature-insensitive, Mistral-7B FP16 stays near $34$--$35\%$ across the full range and SmolLM3-3B INT4 drifts down only slightly ($44.1\%\to37.9\%$). The per-benchmark decomposition in Figure~\ref{fig:bench_temp} shows that this flatness arises from \emph{two distinct mechanisms}, not one. For Mistral-7B the benchmarks move in opposite directions and cancel: AdvBench and GCG-AdvBench ASR rise by $+14.5$\,pp from $T{=}0$ to $1.0$ while ManyHarm falls by $-26$\,pp, leaving the macro-average nearly constant, genuine cancellation. For the saturated models (Granite, SmolLM3-3B) the mean is flat for a different reason: each benchmark is individually near-flat (ManyHarm pinned at $\approx$100\%, the easy sets near $0\%$), so there is little to cancel. The shared conclusion is that temperature's real effect is on decision \emph{stability}, not the pooled mean.

The least stable models are Mistral-7B and SmolLM3-3B (FP16/$T{=}1.0$ DFR $=41.9\%$ and $38.6\%$), while Llama-3.2-3B, Granite-3.1-2B, and Llama-3.1-8B are the most stable (DFR $\leq 5.6\%$). The two Qwen variants, near-perfectly stable when measured on AdvBench alone, show moderate pooled instability (DFR $\approx 15\%$) once the harder benchmarks are included. Over-refusal rate (ORR) stays below 8\% for six of eight models, but the two Qwen variants over-refuse the benign probe set (Qwen3-4B up to 21.5\%), indicating that for the most conservative models usability, rather than attack success, is the dominant deployment cost.

\subsection{Compound Degradation Index}
\label{sec:cdi}

Figure~\ref{fig:cdi} shows CDI heatmaps for all eight models. The key findings:

\begin{figure*}[t]
  \centering
  \includegraphics[width=\linewidth]{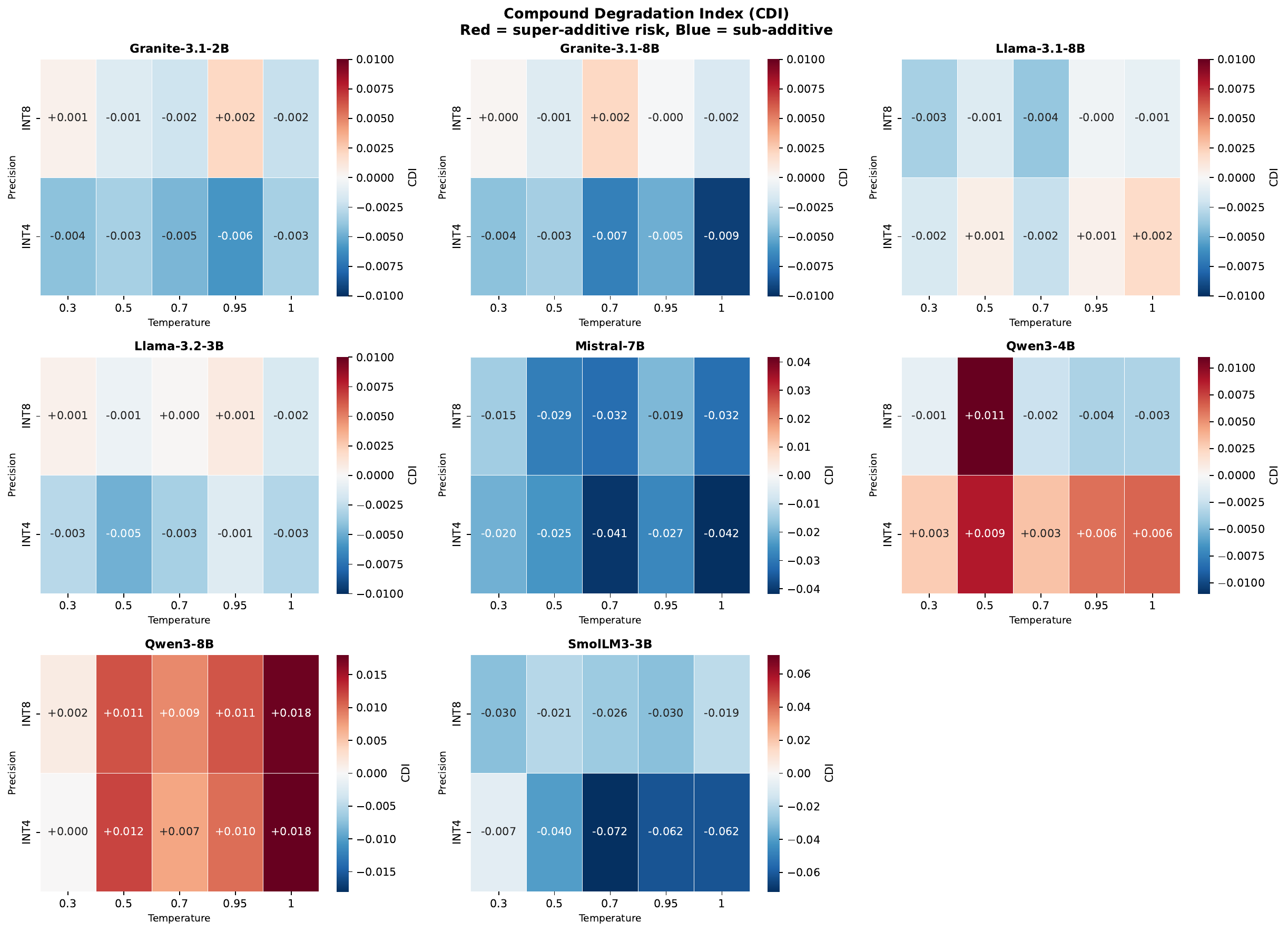}
  \caption{Compound Degradation Index heatmaps (precision $\times$ temperature per model, pooled over the seven benchmarks). Red = super-additive (compounding); blue = sub-additive (countervailing). Most cells cluster near zero.}
  \label{fig:cdi}
\end{figure*}

\textbf{Predominantly sub-additive.} Pooled CDI values range from $-0.071$ to $+0.018$. No model exhibits compounding behavior of any meaningful magnitude, and the majority of CDI values cluster within $\pm0.03$.

\textbf{Strongest sub-additivity (SmolLM3-3B INT4).} CDI reaches $-0.071$ at $T{=}0.7$ (CI $[-0.097, -0.047]$), indicating that at elevated temperatures, INT4 quantization partially \emph{offsets} temperature-induced safety degradation in this model, a countervailing interaction.

\textbf{Mistral-7B sub-additivity.} Both INT4 and INT8 CDI values for Mistral-7B are consistently sub-additive, reaching $-0.042$ (INT4) and $-0.032$ (INT8), so quantization and temperature effects counteract rather than compound.

\textbf{No meaningful super-additivity.} The largest positive CDI across the entire study is $+0.018$ (Qwen3-8B INT4 at $T{=}1.0$; CI $[+0.005, +0.031]$). This interaction is statistically resolvable but practically negligible, a $<2$\,pp effect. Across all 80 interaction cells, 20 have a 95\% CI excluding zero, but every one is small in magnitude ($|$CDI$|\leq0.071$); no configuration shows a positive compounding effect of practical size.

\textbf{Near-independence for the best-aligned models.} Only Llama-3.1-8B and Llama-3.2-3B show CDI near zero ($|$CDI$| \leq 0.005$) across all temperatures; the higher-ASR models (Granite, Mistral, SmolLM) exhibit larger, but still uniformly sub-additive, interactions, consistent with quantization slightly damping temperature-induced degradation rather than amplifying it.

\subsection{Safety Stability}

\begin{figure*}[t]
  \centering
  \includegraphics[width=\linewidth]{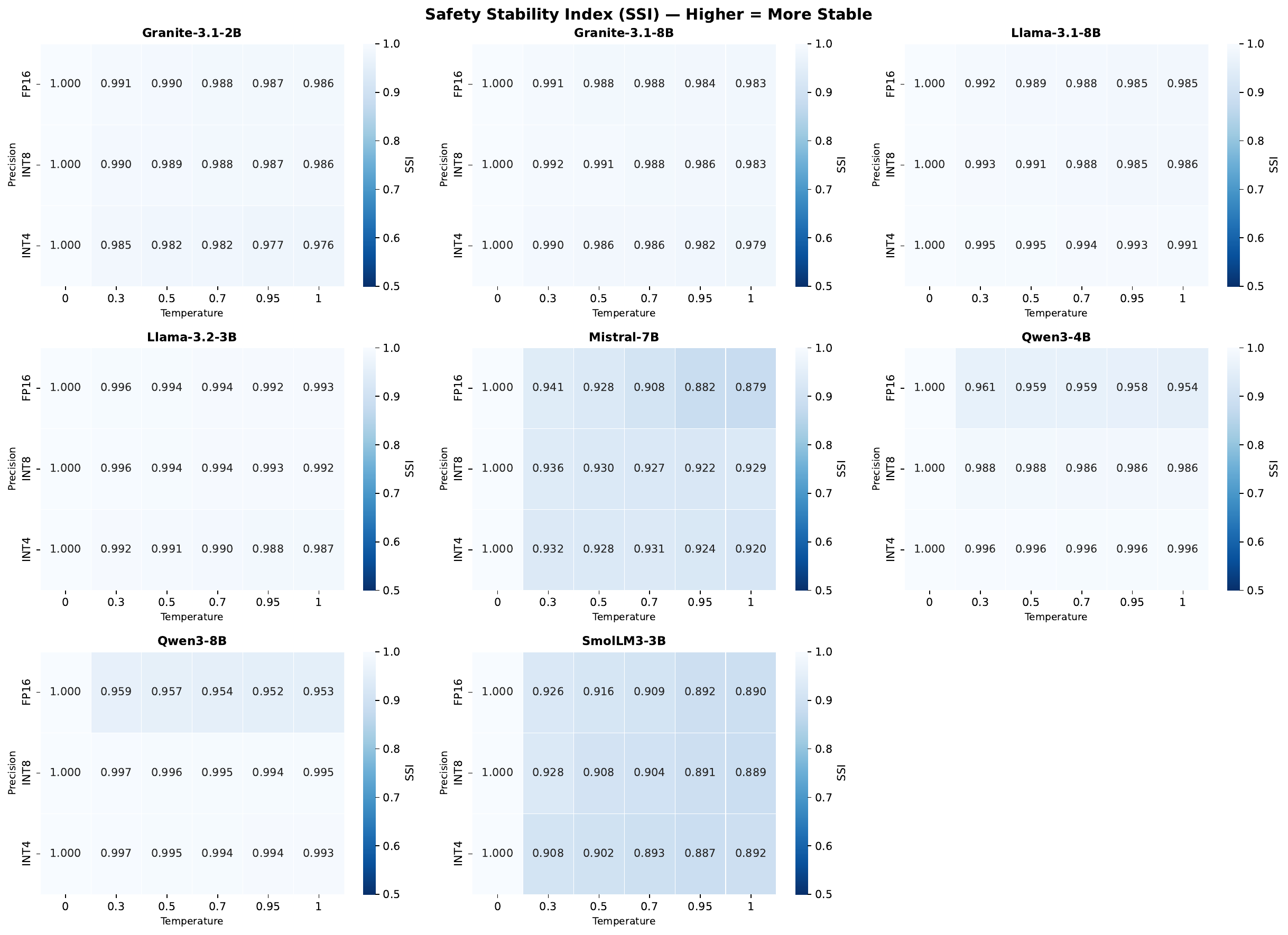}
  \caption{Safety Stability Index (SSI) heatmaps per model (pooled over the seven benchmarks). Stability degrades sharply with temperature for Mistral-7B and SmolLM3-3B, moderately for the two Qwen models, and remains near 1.0 for the rest. Quantization contributes minimally.}
  \label{fig:ssi}
\end{figure*}

Figure~\ref{fig:ssi} shows SSI heatmaps. At $T{=}0$ (greedy) SSI $= 1.0$ for all configurations by definition (deterministic decoding). As temperature increases, stability degrades sharply for two models, moderately for the two Qwen variants, and stays high for the rest:

\begin{itemize}[leftmargin=*, topsep=2pt, itemsep=1pt]
  \item Mistral-7B FP16/$T{=}1.0$: SSI $= 0.879$; DFR $= 41.9\%$ [$39.8, 44.0$] (highest instability).
  \item SmolLM3-3B FP16/$T{=}1.0$: SSI $= 0.890$; DFR $= 38.6\%$ [$36.4, 40.7$].
  \item Qwen3-4B FP16/$T{=}1.0$: SSI $= 0.954$; DFR $= 15.6\%$ [$14.3, 16.9$].
  \item Qwen3-8B FP16/$T{=}1.0$: SSI $= 0.953$; DFR $= 15.4\%$ [$14.3, 16.6$].
  \item Llama-3.1-8B FP16/$T{=}1.0$: SSI $= 0.985$; DFR $= 5.6\%$ [$4.5, 6.7$].
  \item Llama-3.2-3B FP16/$T{=}1.0$: SSI $= 0.993$; DFR $= 2.8\%$ [$2.1, 3.5$] (most stable).
\end{itemize}
(Bracketed ranges are 95\% bootstrap CIs; the DFR ordering is well separated and robust.)

\textbf{Temperature dominates.} Quantization alone causes only small SSI change ($<0.02$) at any fixed temperature, whereas temperature is the dominant instability driver. For Mistral-7B and SmolLM3-3B, single-sample safety evaluations at $T{>}0.7$ are unreliable, with FP16 DFR reaching 41.9\% and 38.6\% respectively at $T{=}1.0$, the study-wide maximum pooled DFR is 41.9\% (Mistral-7B FP16). Notably, the two Qwen models, which appear perfectly stable on AdvBench alone, reach DFR $\approx 15\%$ once the harder benchmarks are pooled in, so single-benchmark stability estimates can also be over-optimistic.

\subsection{Alignment Score and Over-Refusal}

The alignment score, defined as $\text{RR} - \text{ORR}$, captures the net safety benefit accounting for over-restriction costs. Only the two Llama models score $\geq 0.97$ at FP16/$T{=}0$ (Llama-3.2-3B $0.989$, Llama-3.1-8B $0.979$). The lowest scores, SmolLM3-3B ($0.650$), Qwen3-4B ($0.658$), and Mistral-7B ($0.661$), arise from different failure modes: Mistral-7B and SmolLM3-3B from high pooled ASR ($33.9\%$, $34.5\%$) with near-zero over-refusal, whereas Qwen3-4B is penalised on \emph{both} axes, combining $14.2\%$ pooled ASR with $20.0\%$ over-refusal of benign prompts. SmolLM3-3B's score falls further under INT4 (to $0.544$), reflecting its quantization sensitivity. ORR remains below 8\% for six of eight models; only the Qwen variants over-refuse substantially (Qwen3-4B up to 21.5\% at INT8/$T{=}0$, Qwen3-8B 8.0\% at FP16/$T{=}0$).

\subsection{Quantization as a Constant ASR Offset}
\label{sec:equiv}

A natural deployment question is whether quantization can be summarized as an equivalent change in sampling temperature, i.e.\ whether a quantized model behaves like the FP16 model run at some shifted $T^*$. Our data answer this in the negative: because pooled ASR is nearly temperature-independent (Figure~\ref{fig:asr_temperature}), no temperature shift reproduces the quantization effect. Instead, quantization is well described as a near-constant, temperature-independent ASR \emph{offset} relative to FP16. This is the more useful summary, and Table~\ref{tab:equiv} reports it.

\begin{table}[t]
\centering
\caption{Quantization summarized as a pooled ASR offset relative to FP16 (six-judge ensemble), at $T{=}0$. The offset is near-constant across temperature (Table~\ref{tab:delta} gives the $T{=}0$ values with CIs).}
\label{tab:equiv}
\small
\begin{tabular}{lll}
\toprule
\textbf{Model group} & \textbf{Prec.} & \textbf{$\Delta$ASR vs FP16} \\
\midrule
7 of 8 models            & INT8 / INT4 & within $\pm1.6$\,pp \\
SmolLM3-3B               & AWQ INT4    & $+9.6$\,pp (riskier) \\
\bottomrule
\end{tabular}
\end{table}

For seven of the eight models, and for GPTQ INT8 across the board, the quantization offset is small ($\leq 1.6$\,pp) and does not vary systematically with temperature. SmolLM3-3B is the sole model where INT4 raises pooled ASR materially ($+9.6$\,pp), making it consistently riskier than FP16.

\subsection{Judge Agreement Analysis}
\label{sec:judge_agreement}

We evaluate all 144 configurations with six safety judges and compare verdicts at the individual response level ($\approx$367k harmful prompt-level entries, 5 samples each).

\begin{table}[t]
\centering
\caption{Cross-judge config-level ASR agreement (within 5\,pp of primary judge LlamaGuard-3-1B) and mean signed ASR difference across 144 evaluated configurations.}
\label{tab:judge}
\small
\begin{tabular}{lcc}
\toprule
\textbf{Judge} & \textbf{Config Agree.} & \textbf{Mean $\Delta$ ASR} \\
\midrule
LlamaGuard-3-8B   & 93.1\% & $+$0.007 \\
LlamaGuard-2-8B   & 78.5\% & $-$0.026 \\
WildGuard         & 33.3\% & $-$0.103 \\
ShieldGemma-2B    & 75.0\% & $-$0.038 \\
ShieldGemma-9B    & 100.0\% & $-$0.014 \\
\bottomrule
\end{tabular}
\end{table}

Cross-judge agreement is substantially lower on the pooled multi-benchmark suite than on AdvBench alone, which underscores the value of the ensemble. ShieldGemma-9B and LlamaGuard-3-8B track the primary judge (LlamaGuard-3-1B) closely (agreement $100\%$ and $93.1\%$; mean $\Delta = -0.014$ and $+0.007$), whereas WildGuard diverges sharply, reading much lower ASR (mean $\Delta = -0.103$, agreement only $33.3\%$); LlamaGuard-2-8B and ShieldGemma-2B sit in between ($-0.026$, $-0.038$). The disagreement is concentrated on the harder benchmarks (HarmBench, ManyHarm) and on borderline prompts for Mistral-7B and SmolLM3-3B, consistent with prior work \cite{dubey2024llama3herdmodels}. The six-judge majority-vote ensemble used throughout this paper mitigates these individual judge biases.

\subsection{Cross-Benchmark Generalization}
\label{sec:crossbench}

\begin{figure*}[t]
  \centering
  \includegraphics[width=0.82\linewidth]{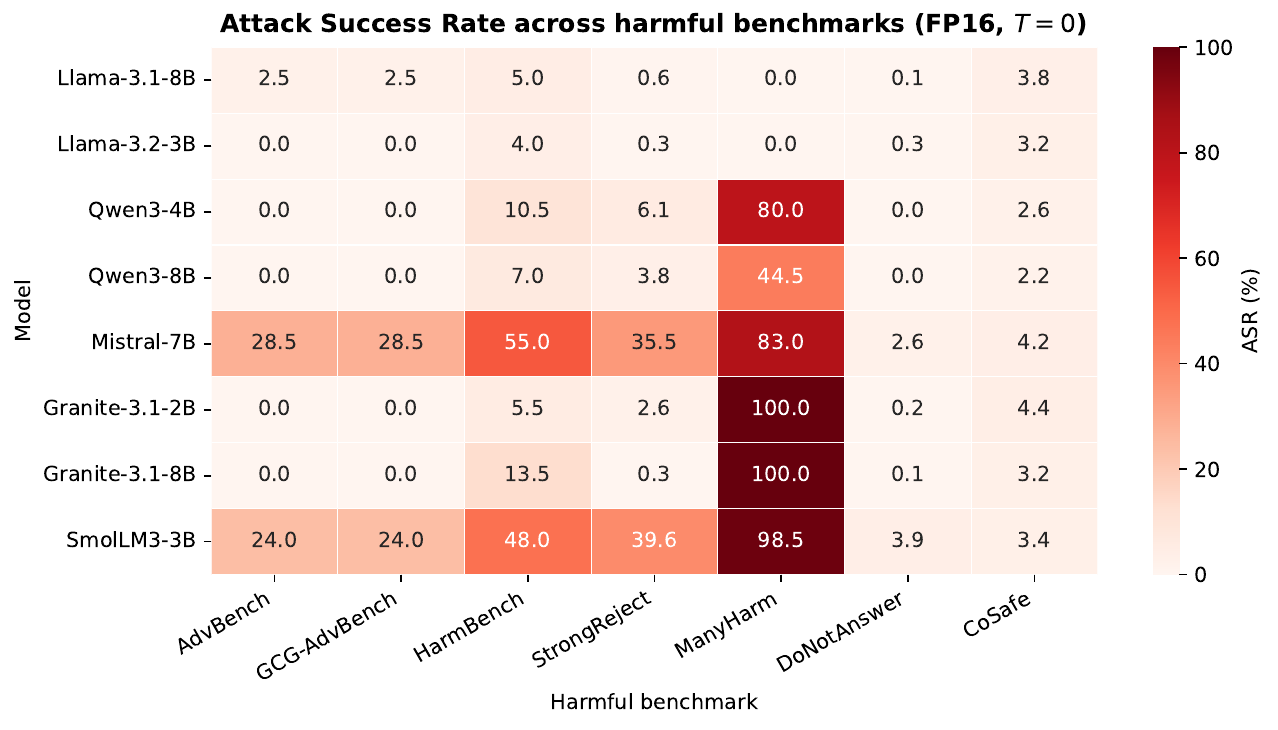}
  \caption{Attack Success Rate (FP16, $T{=}0$, six-judge ensemble) for each model across seven harmful benchmarks. Models that appear fully aligned on AdvBench (e.g.\ both Granite variants, Qwen3-4B at 0.0\%) are highly vulnerable on ManyHarm (80--100\%) and elevated on HarmBench/StrongREJECT, showing that AdvBench-only evaluation understates absolute risk.}
  \label{fig:dataset_asr}
\end{figure*}

All metrics above are macro-averaged over the seven harmful benchmarks. Figure~\ref{fig:dataset_asr} decomposes that pooled ASR into its per-benchmark, per-model components (FP16/$T{=}0$, same six-judge ensemble), showing where the risk concentrates and how misleading any single benchmark can be.

\textbf{AdvBench understates absolute risk.} Mean ASR across the 8 models is 6.9\% on AdvBench (identical on GCG-AdvBench) but climbs to 18.6\% on HarmBench and 63.2\% on ManyHarm. The gap is most dramatic for models that look perfectly aligned on AdvBench: Granite-3.1-2B and Granite-3.1-8B jump from 0.0\% on AdvBench to 100\% on ManyHarm, Qwen3-4B from 0.0\% to 80.0\%, and SmolLM3-3B from 24.0\% to 98.5\%. DoNotAnswer (0.9\%) and CoSafe (3.4\%) remain easy for every model.

\textbf{The absolute ManyHarm level is judge-sensitive.} ManyHarm is also where the six judges disagree most, so its \emph{absolute} ASR should be read with care. At FP16/$T{=}0$ the per-judge ManyHarm ASR spans a wide range, e.g.\ $3.5$--$100\%$ for Granite-3.1-8B, $10.5$--$86\%$ for Mistral-7B, $22.5$--$99\%$ for SmolLM3-3B, and $19$--$95.5\%$ for Qwen3-4B, and the six-judge ensemble sits near the top of that range. The \emph{qualitative} finding, that AdvBench-safe models are not safe under many-shot framing, holds for the ensemble and for the majority of individual judges, but the precise absolute number on this benchmark is a property of the judge ensemble as much as of the target model, which is exactly why we report a majority vote of six judges rather than any single one. ManyHarm measures susceptibility to many-shot \emph{adversarial framing}, not baseline single-turn alignment; we therefore read it as an upper-tail stress test rather than a deployment ASR.

\textbf{Relative ordering only partially transfers.} The two least-safe models on AdvBench, Mistral-7B and SmolLM3-3B, stay among the least safe on every benchmark, and the easy benchmarks are uniformly easy. But the AdvBench-safe models do \emph{not} stay safe: their risk is benchmark-dependent and surfaces under prompt distributions, especially ManyHarm's repeated-harm framing, that AdvBench does not exercise. A single-benchmark safety verdict can therefore be badly miscalibrated.

\textbf{Implication for evaluation.} Because the pooled risk is dominated by a few benchmarks, the choice of benchmark dominates any absolute safety claim. The quantization and temperature effects in Sections~\ref{sec:baseline}--\ref{sec:judge_agreement} are \emph{relative} shifts that hold across the pooled suite, but the \emph{absolute} safety of a model should always be read across benchmarks, never from AdvBench (or any single set) alone, a model at $0\%$ on AdvBench can exceed $80\%$ on ManyHarm.

%% ─────────────────────────────────────────────
\section{Discussion}
\label{sec:discussion}

\paragraph{Standard quantization largely preserves safety.}
Across 8 models and two quantization methods, AWQ INT4 stays within $\approx$1.6\,pp of pooled FP16 ASR (or reduces it) for seven of the eight models, and GPTQ INT8 is similarly close. SmolLM3-3B is the clear exception, where INT4 raises pooled ASR from $34.5\%$ to $44.1\%$. This suggests that models with strong baseline alignment are robust to standard post-training quantization, while models with weaker alignment are more vulnerable to quantization-induced safety degradation. This pattern is consistent with prior observations on Llama models \cite{chen2025q}, but our broader model coverage shows it is not universal. A separate, more actionable failure mode is that the default \texttt{llmcompressor} AWQ export silently \emph{collapsed} into degenerate outputs for two of eight models (Qwen3-4B, Llama-3.2-3B), recoverable only with a dedicated re-quantization run; standard quantization pipelines can thus produce broken, not merely less-safe, checkpoints, so a collapse/quality gate should precede any safety claim.

\paragraph{Temperature is the dominant risk factor.}
While quantization effects on pooled ASR are small and model-dependent, temperature has a direct effect on decision \emph{instability}. At FP16/$T{=}1.0$, pooled DFR ranges from $2.8\%$ (Llama-3.2-3B) to $41.9\%$ (Mistral-7B) and $38.6\%$ (SmolLM3-3B), with the two Qwen variants intermediate ($\approx$15\%). Instability is thus concentrated in the weaker-aligned models rather than being a universal phenomenon, but the Qwen case shows that AdvBench-only stability estimates (where both Qwen models read DFR $\approx 0$) can be over-optimistic.

\paragraph{Compound effects do not compound.}
The CDI analysis reveals that practitioners need not fear a ``double penalty'' from combining quantization and high temperature. Pooled CDI values range from $-0.071$ to $+0.018$, with the majority clustering near zero. The strongest effects are \emph{sub-additive}: SmolLM3-3B and Mistral-7B reach CDI of $-0.071$ and $-0.042$ respectively, indicating that quantization partially offsets temperature-induced ASR degradation for these models. No configuration shows super-additivity of any practical size, the largest positive CDI in the entire study is $+0.018$ (CI $[+0.005,+0.031]$), a statistically resolvable but practically negligible $<2$\,pp interaction.

\paragraph{Limitations.}
Our study uses AWQ (INT4) and GPTQ (INT8) with standard Pile validation calibration data only; NF4/BitsAndBytes, GGUF, or adversarial calibration strategies may behave differently. Two INT4 configurations (Qwen3-4B, Llama-3.2-3B) required a dedicated AWQ re-quantization run after the default \texttt{llmcompressor} export collapsed into degenerate outputs, so deployment-grade quantization should validate each checkpoint for collapse before use. Pooled ASR is macro-averaged over the seven benchmarks; a different weighting (e.g.\ micro-averaging by prompt count) shifts the absolute numbers, though the relative quantization and temperature effects are robust to this choice. The benign over-refusal set is a 200-prompt XSTest-style probe. We evaluate 2B--8B models; findings may not generalize to models $>$70B, and all prompt sets are static, so adaptive jailbreaks may show different patterns. Safety judges are themselves imperfect, as our agreement analysis shows.

%% ─────────────────────────────────────────────
\section{Conclusion}
\label{sec:conclusion}

We conducted a broad empirical study of how two standard deployment choices, post-training quantization and sampling temperature, jointly affect LLM safety alignment across 144 configurations spanning 8 models, 3 precisions (FP16, GPTQ INT8, AWQ INT4), and 6 temperatures, evaluated over a seven-benchmark harmful suite on 8$\times$NVIDIA H100 GPUs with a six-judge safety ensemble. All quantized models use standard calibration data (Pile validation set) without adversarial manipulation, reflecting realistic deployment pipelines. Our headline findings: (1)~standard quantization is approximately safety-neutral, with AWQ INT4 tracking pooled FP16 ASR within $\approx$1.6\,pp for seven of eight models, only SmolLM3-3B degrading substantially ($34.5\%\to44.1\%$); (2)~temperature is the dominant instability driver, concentrated in Mistral-7B (DFR $= 41.9\%$) and SmolLM3-3B (DFR $= 38.6\%$ at FP16/$T{=}1.0$) while the best-aligned models stay below $6\%$; and (3)~the quantization$\times$temperature interaction is predominantly sub-additive ($-0.071 \leq \text{CDI} \leq +0.018$), with quantization partially offsetting temperature effects rather than compounding them. Critically, a per-benchmark decomposition shows these pooled numbers still understate worst-case risk, several models score $0\%$ on AdvBench yet exceed $80\%$ on ManyHarm, so we recommend that safety evaluations report multi-sample stability metrics alongside ASR \emph{and} span multiple benchmarks, and that practitioners verify quantization safety specifically for models with weak baseline alignment. The CDI framing and per-benchmark decomposition offer new tools for deployment-aware safety evaluation.

%% ─────────────────────────────────────────────
\balance
\bibliographystyle{unsrt}
\bibliography{refs}

\end{document}